\documentclass[a4paper,fleqn]{cas-dc}
\usepackage[numbers]{natbib}
\usepackage{graphicx}
\usepackage{subcaption}
\usepackage{booktabs}
\usepackage{multirow}
\usepackage{float}

\begin{document}
\let\WriteBookmarks\relax
\def\floatpagepagefraction{1}
\def\textpagefraction{.001}
\let\printorcid\relax 

% \shorttitle{<short title of the paper for running head>} 
\shorttitle{}    

% \shortauthors{<short author list for running head>}
\shortauthors{Zhicheng Wang et al.}

%论文标题
\title[mode = title]{A Low-Cost Real-Time Framework for Industrial Action Recognition Using Foundation Models}  

%作者信息
\author[1]{Zhicheng Wang}
\fnmark[1]
\author[2]{Wensheng Liang}
\fnmark[1]
\author[3]{Ruiyan Zhuang}
\author[4]{Shuai Li}
\author[1]{Jianwei Tan}
\author[4]{Xiaoguang Ma}
\cormark[1]

\address[1]{College of Information Science and Engineering, Northeastern University, Shenyang, China} 
\address[2]{School of Mechanical Engineering and Automation, Northeastern University, Shenyang, China} 
\address[3]{Enterprise AI, Midea AI Innovation Center, Foshan, China}
\address[4]{Foshan Graduate School of innovation, Northeastern University, Foshan, China} 
\cortext[4]{\textsuperscript{*}Corresponding Author.}  
\cortext[4]{\textsuperscript{1} These authors contributed equally to this work.}  
\cortext[4]{\textit{E-mail addresses:} \href{mailto:2390108@stu.neu.edu.cn}{2390108@stu.neu.edu.cn} (Z. Wang), \href{mailto: 2200385@stu.neu.edu.cn}{ 2200385@stu.neu.edu.cn} (W. Liang), \href{mailto: zhuangry@midea.com}{zhuangry@midea.com} (R. Zhuang), \href{mailto: shuaili1027@gmail.com}{shuaili1027@gmail.com} (S. Li), \href{mailto: 2410359@stu.neu.edu.cn}{2410359@stu.neu.edu.cn} (J. Tan), \href{mailto: maxg@mail.neu.edu.cn}{maxg@mail.neu.edu.cn} (X. Ma).}  
\cortext[4]{\textsuperscript{\dag} The paper is under consideration at Pattern Recognition Letters.}  
\begin{abstract}
Action recognition (AR) in industrial environments---particularly for identifying actions and operational gestures---faces persistent challenges due to high deployment costs, poor cross-scenario generalization, and limited real-time performance.
To address these issues, we propose a low-cost real-time framework for industrial action recognition using foundation models, denoted as LRIAR, to enhance recognition accuracy and transferability while minimizing human annotation and computational overhead. 
The proposed framework constructs an automatically labeled dataset by coupling Grounding DINO with the pretrained BLIP-2 image encoder, enabling efficient and scalable action labeling.
Leveraging the constructed dataset, we train YOLOv5 for real-time action detection, and a Vision Transformer (ViT) classifier is deceloped via LoRA-based fine-tuning for action classification.
Extensive experiments conducted in real-world industrial settings validate the effectiveness of LRIAR, demonstrating consistent improvements over state-of-the-art methods in recognition accuracy, scenario generalization, and deployment efficiency.  
\end{abstract}
%	\item A low-cost real-time framework (LRIAR) for industrial action recognition is proposed, leveraging large-scale foundation models to enhance recognition accuracy and cross-scenario generalization.
%	
%	\item Grounding DINO and BLIP-2 are integrated to automatically construct high-quality industrial action datasets, significantly reducing human annotation costs.
%	
%	\item YOLOv5 is used for real-time action detection, and Vision Transformer (ViT) is fine-tuned with LoRA for lightweight classification, improving deployment efficiency.
%	
%	\item LRIAR demonstrates superior cross-line generalization and real-time performance in real-world industrial settings, outperforming state-of-the-art methods.
%	
%	\item Extensive experiments validate the framework’s significant improvements in accuracy, deployment efficiency, and annotation costs, making it highly applicable for smart manufacturing.

% \begin{highlights}
% \item highlight-1
% \item highlight-2
% \item highlight-3
% \end{highlights}

% Keywords
\maketitle
\section{Introduction}
With the rapid advancement of intelligent manufacturing, action recognition (AR) has emerged as a key technology in modern production systems \cite{sun2022human}.
In particular, AR plays a critical role in tasks such as human-machine interaction and worker behavior monitoring, supporting real-time decision-making in quality control, safety management, and process optimization \cite{wang2024multi,huang2025rcst}.
However, despite their success on open-domain benchmarks, current state-of-the-art AR models often exhibit degraded performance in real-world industrial settings \cite{8531764,9209531}. 

This performance gap primarily stems from two challenges: the lack of domain-specific training data tailored to industrial environments and the substantial variability across manufacturing lines, which leads to significant domain shifts and poor cross-scenario generalization \cite{10195224,jiang2024swarm}.
As a specialized branch of AR, industrial human action recognition (IHAR) faces additional challenges. 

Traditional IHAR approaches typically combine object detectors (e.g., YOLOv5 \cite{glenn_jocher_2022_7347926}) with image classifiers (e.g., ResNet-18 \cite{He_2016_CVPR}), relying heavily on manually annotated data for supervised training \cite{WANG2024110527}. 
Although these methods offer moderate computational efficiency, their dependence on large-scale manual labeling results in high labor costs and long development cycles \cite{WANG2024710}. 
More critically, they often fail to meet the real-time and low-deployment-cost requirements of industrial applications \cite{10726628}.
\begin{figure}
	\centering
	\centerline{\includegraphics[width=3.5in]{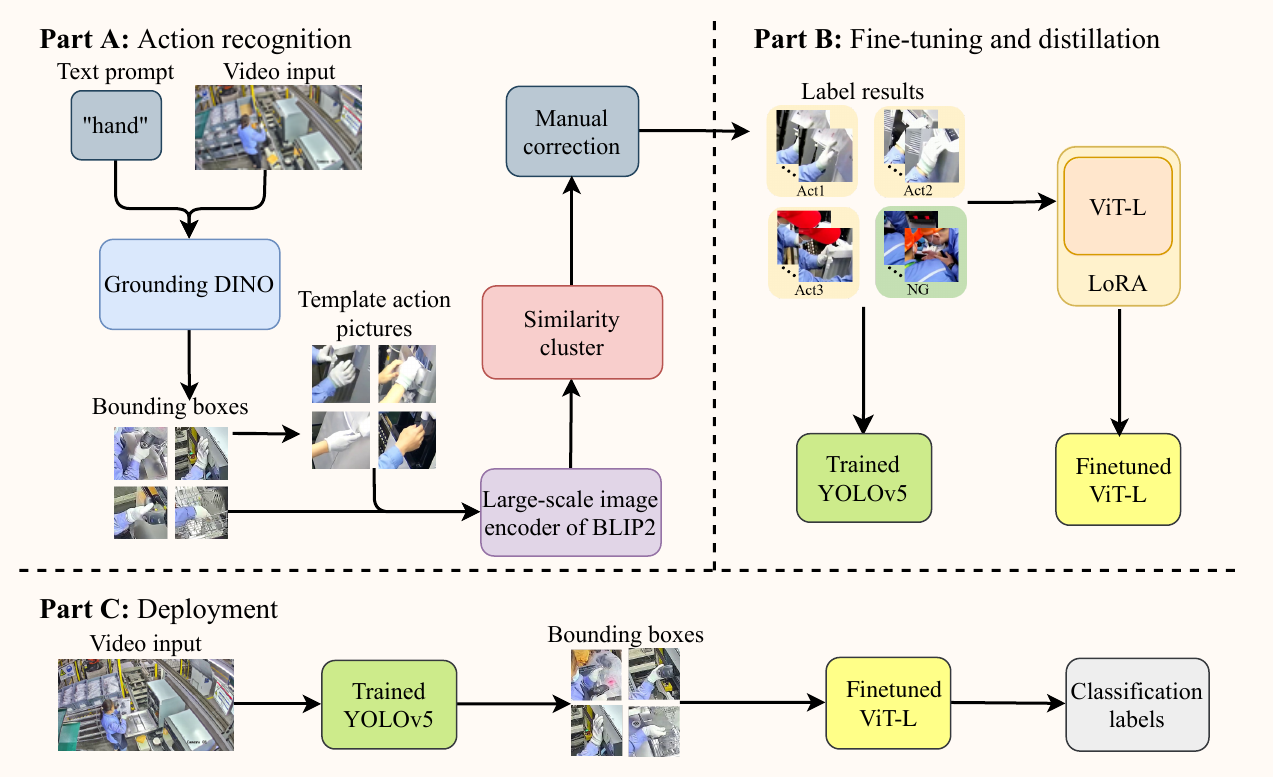}}
	\caption{Overview of the proposed low-cost real-time industrial action recognition framework.}\label{f1}
\end{figure}

Recently, large-scale foundation models (LSFMs) have shown promising potential in visual and language understanding tasks, offering strong generalization and cross-modal learning capabilities \cite{10922728}. 
Pretrained on massive heterogeneous datasets, LSFMs can capture complex spatiotemporal and semantic patterns, and perform automatic annotation via vision-language alignment, significantly reducing human annotation costs \cite{zhou2024towards,zhang2025large}. 
However, their high inference latency and computational demands hinder direct deployment in resource-constrained industrial scenarios \cite{10478189}.

To address these limitations, we propose a novel Low-cost Real-time Industrial Action Recognition (LRIAR) framework that leverages LSFMs for efficient, scalable IHAR.
Our approach integrates Grounding DINO \cite{liu2023grounding} and BLIP-2 \cite{a3619222} to automatically extract action-relevant bounding boxes and generate semantic labels, minimizing human supervision. 
The resulting pseudo-labeled samples are used to train a YOLOv5-based action detector. Furthermore, we apply Low-Rank Adaptation (LoRA) \cite{liu2024grounding} to fine-tune a Vision Transformer (ViT) \cite{9716741}, yielding a lightweight and high-performance classifier \cite{jiang2024leveraging}. 
Experimental results demonstrate that LRIAR achieves superior cross-line generalization and real-time performance, while significantly reducing annotation and deployment costs—making it well-suited for practical industrial use.
\section{Related work}
Industrial human action recognition (IHAR) plays a pivotal role in intelligent manufacturing, particularly in applications such as worker behavior monitoring, quality control, and process optimization \cite{wang2024multi,shi2025frefusion}.
However, existing IHAR methods face several challenges, primarily involving high annotation costs, poor cross-scenario generalization, and real-time performance issues.
This paper focuses on addressing these challenges by investigating methods for cost-effective construction of high-quality industrial datasets, while ensuring dataset stability and accuracy.

\subsection{Challenges in Constructing Industrial AR Datasets}
Constructing high-quality industrial action recognition datasets typically requires extensive manual annotation, resulting in significant data collection costs. Unlike large-scale video datasets used in non-industrial contexts (e.g., UCF101 \cite{soomro2012ucf101} and Kinetics \cite{carreira2017quo}), industrial environments demand the collection of data specific to particular production lines and task categories, making dataset construction more complex and expensive \cite{liu2021no, vieira2022low}. 
Recent studies have primarily focused on reducing manual intervention through automated annotation methods. However, due to the diversity of industrial scenarios, existing methods often struggle to adapt effectively to different production lines and tasks \cite{WANG2024710}.

To address this issue, recent research has explored methods such as semi-supervised learning and self-supervised learning, which automatically annotate data using a small number of labeled samples, thereby reducing manual labor \cite{8006248,huang2025t}. Additionally, the use of transfer learning and pretrained models has allowed models trained in non-industrial environments to be transferred to industrial tasks, reducing the need for labeled data \cite{cao2020automatic}. While these methods have made progress in reducing annotation costs, they often fail to ensure the accuracy and stability of industrial datasets, particularly when dealing with the complexity of production line operations.

\subsection{Application of Large-Scale Foundation Models in Industrial Dataset Construction}
With the development of Large-Scale Foundation Models (LSFM), such as CLIP \cite{radford2021learning}, Grounding DINO \cite{liu2023grounding}, and BLIP-2 \cite{li2023blip}, the efficiency and accuracy of automated annotation have been significantly improved. LSFMs, through vision-language alignment, can automatically detect and label target regions in images, dramatically reducing the reliance on manual annotation \cite{zhou2024towards, zhang2025large,gu2024large}. For instance, Grounding DINO utilizes a dual-modality visual and textual encoder to effectively detect regions of interest in industrial images, enabling precise automated annotation \cite{liu2023grounding}.

However, despite the impressive performance of LSFMs in annotation tasks, their high computational complexity and inference latency still limit their applicability in real-time industrial settings. To overcome these challenges, researchers have proposed techniques such as model lightweighting and Low-Rank Adaptation (LoRA), which reduce model parameters and accelerate inference, thereby improving real-time performance \cite{hu2021lora,liu2024grounding}. While these methods effectively enhance inference speed, they still face challenges in completely addressing the needs of industrial applications, especially in terms of reducing computational resource consumption without compromising annotation accuracy.
\begin{figure}
	\centering
	\centerline{\includegraphics[width=3.5in]{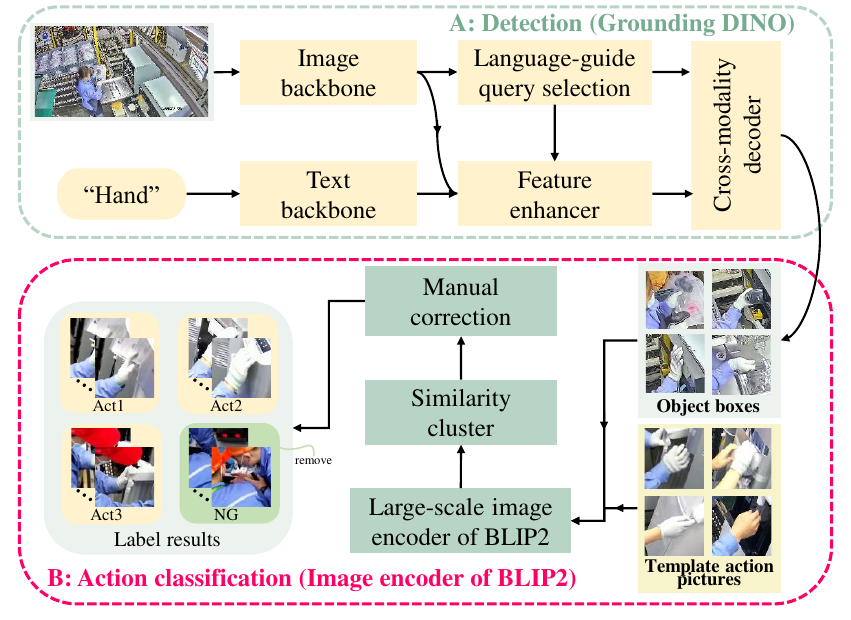}}
	\vspace{-6pt}
	\caption{Detected processes of the industrial action recognition.}\label{f2}
\end{figure}
\section{Methodology}
We propose a three-stage framework for Low-cost Real-time Industrial Action Recognition (LRIAR), consisting of: 1) automatic construction of a high-quality action dataset using large-scale foundation models (LSFMs); 2) collaborative training of an action detector and a lightweight action classifier; and 3) deployment in real-time industrial scenarios. The overall pipeline is illustrated in Fig.~\ref{f1}.
\subsection{Dataset Construction via Large-Scale Foundation Models}\label{3.1} 
To address the high cost of manual annotations in industrial environments, we design an automated labeling strategy that integrates semantic-guided detection and image-level semantic matching, thereby minimizing human effort while preserving label quality.
\subsubsection{Semantic-Guided Action Detection}
Let $\{I_t\}$ denote a sequence of video frames and $T$ be a predefined semantic prompt (e.g., ``hand''). We employ Grounding DINO, a zero-shot object detector, to extract candidate action regions $B_t$ from each frame:
\begin{equation}
B_t = \mathcal{D}_{\text{GD}}(I_t, T),
\end{equation}
where $\mathcal{D}_{\text{GD}}(\cdot)$ denotes the detection function of Grounding DINO, and $B_t = \{R_i\}_{i=1}^{N_t}$ represents the set of bounding boxes in frame $I_t$.

Grounding DINO employs a dual-encoder architecture and cross-modal decoder, wherein image and text embeddings are independently encoded and subsequently fused via a multi-stage feature enhancer.
A language-guided query mechanism is then utilized to generate decoder inputs, resulting in accurate and semantically grounded region proposals, as shown in Fig.~\ref{f2} (A).
\subsubsection{Semantic Matching and Label Assignment via BLIP-2}
To assign semantic labels to the detected regions, we adopt a pretrained BILP-2 model with a frozen image encoder \cite{wu2024hand}.
For each candidate region $R_i \in B_t$, we compute the cosine similarity with a set of $K$ action templates ${T_k}_{k=1}^K$ as:
\begin{equation}
S_{ik} = \frac{\psi(R_i) \cdot \psi(T_k)}{\|\psi(R_i)\| \cdot \|\psi(T_k)\|},
\end{equation}
where $\psi(\cdot)$ is the BLIP-2 image encoder and $S_{ik}$ denotes the similarity score between region $R_i$ and template $T_k$. Label assignment is then performed as follows:
\begin{equation}
\hat{y}_i =
\begin{cases}
\arg\max_k S_{ik}, & \text{if } \max_k S_{ik} \geq \lambda, \\
\text{NG}, & \text{otherwise},
\end{cases}
\end{equation}
where $\lambda$ is a predefined similarity threshold and NG represents invalid or unmatched regions, as depicted in Fig.~\ref{f2} (B).
\subsubsection{Lightweight Human Verification}
In practice, the automated labeling process achieves an accuracy exceeding 97\%. 
To further improve label quality, a lightweight manual verification step is applied to ambiguous or low-confidence samples. 
The final training dataset is denoted as:
\begin{equation}
\mathcal{D}_{L} = \{(R_i, \hat{y}_i)\}_{i=1}^N.
\end{equation}
\subsection{Joint Training of Detection and Classification Models}
Using the constructed dataset $\mathcal{D}_L$, we jointly train an object detector and a visual classifier to achieve a balance between recognition performance and computational efficiency.

\subsubsection{YOLOv5-Based Action Detection}
We adopt YOLOv5 as the detection model $\mathcal{F}_{\text{det}}$, which is trained to predict action-related bounding boxes from raw frames:
\begin{equation}
\{R_i^t\} = \mathcal{F}_{\text{det}}(I_t),
\end{equation}
where $\{R_i^t\}$ denotes the set of detected bounding boxes at time $t$.
\subsubsection{ViT-L Fine-Tuning via LoRA}
\begin{figure}
	\centering
	\centerline{\includegraphics[width=3in]{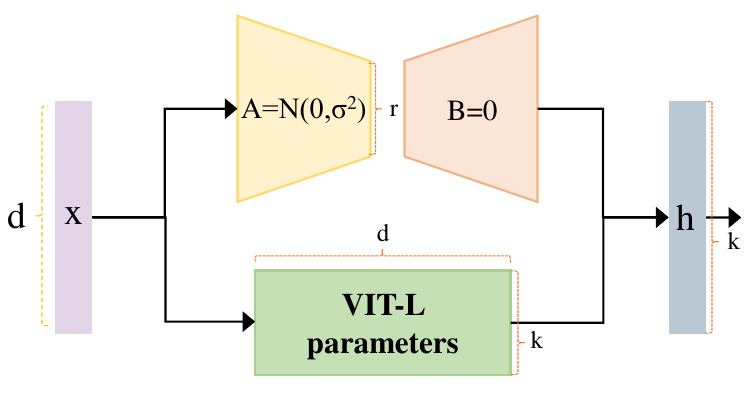}}
	\vspace{-6pt}
	\caption{The LoRA fine-tuning process.}\label{f3}
\end{figure}

To classify detected regions efficiently, we fine-tune a pretrained Vision Transformer (ViT-L) model using Low-Rank Adaptation (LoRA). 
This involves inserting two trainable low-rank matrices $A \in \mathbb{R}^{r \times k}$ and $B \in \mathbb{R}^{d \times r}$ into a frozen weight matrix $W_0$, where $r \ll \min(d, k)$. The adapted transformation is defined as:
\begin{equation}
h = W_0 x + BAx,
\end{equation}
where $x$ is the input embedding and $h$ is the output. Only $A$ and $B$ are updated during training, significantly reducing parameter overhead while preserving model capacity, as illustrated in Fig.~\ref{f3}.
\subsection{Real-Time Deployment in Industrial Environments}
For real-time deployment, we substitute Grounding DINO with the trained YOLOv5 model for efficient region proposal, and utilize the LoRA-adapted ViT-L for classification. The deployed architecture is shown in Fig.~\ref{f1} (C).
At inference time, each frame $I_t$ is processed as:
\begin{equation}
\begin{aligned}
\{R_i^t\} &= \mathcal{F}_{\text{det}}(I_t), \\
\hat{y}_i^t &= \mathcal{F}_{\text{ViT}}(R_i^t),
\end{aligned}
\end{equation}
where $\hat{y}_i^t$ denotes the predicted label for region $R_i^t$.
This hybrid detector-classifier design enables robust, low-latency, and multi-position action recognition across diverse manufacturing stations, facilitating scalable deployment in industrial workflows.
\section{Experiments and Results}

\subsection{Experimental Setup}
\begin{figure}
	\centering
	\begin{subfigure}{0.11\textwidth}
		\includegraphics[width=\linewidth]{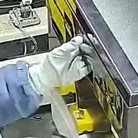}
		\caption{$Act_1$}
		\label{f41}
	\end{subfigure}
	\begin{subfigure}{0.11\textwidth}
		\includegraphics[width=\linewidth]{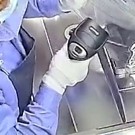}
		\caption{$Act_2$}
		\label{f42}
	\end{subfigure}
	\begin{subfigure}{0.11\textwidth}
		\includegraphics[width=\linewidth]{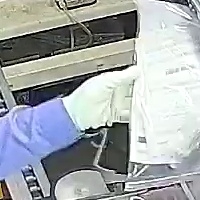}
		\caption{$Act_3$}
		\label{f43}
	\end{subfigure}
	\begin{subfigure}{0.11\textwidth}
		\includegraphics[width=\linewidth]{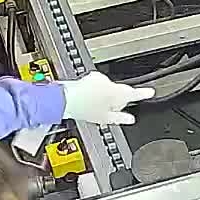}
		\caption{$NG$}
		\label{f44}
	\end{subfigure}
	\caption{Typical actions in GI-D. (a) $Act_1$: door O/C, (b) $Act_2$: check, (c) $Act_3$: instructions insertion, and (d) NG: other irrelevant actions.}
	\label{f4}
\end{figure}

\begin{figure}
	\centering
	\begin{subfigure}{0.09\textwidth}
		\includegraphics[width=\linewidth]{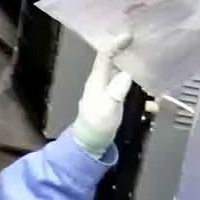}
		\caption{$Act_1$}
		\label{f51}
	\end{subfigure}
	\begin{subfigure}{0.09\textwidth}
		\includegraphics[width=\linewidth]{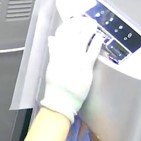}
		\caption{$Act_2$}
		\label{f52}
	\end{subfigure}
	\begin{subfigure}{0.09\textwidth}
		\includegraphics[width=\linewidth]{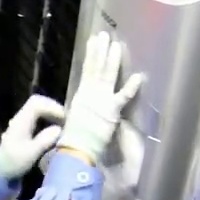}
		\caption{$Act_3$}
		\label{f53}
	\end{subfigure}
	\begin{subfigure}{0.09\textwidth}
		\includegraphics[width=\linewidth]{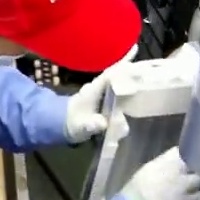}
		\caption{$Act_4$}
		\label{f54}
	\end{subfigure}
	\begin{subfigure}{0.09\textwidth}
		\includegraphics[width=\linewidth]{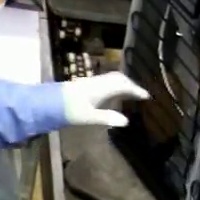}
		\caption{$NG$}
		\label{f55}
	\end{subfigure}
	\caption{Typical actions in GI-WD. (a) $Act_1$: instructions insertion, (b) $Act_2$: top buttons pressing, (c) $Act_3$: logo paste, and (d) $Act_4$: inner check, and (g) $NG$: other irrelevant actions.}
	\label{f5}
\end{figure}

We evaluated the proposed LRIAR framework on two industrial assembly lines at Midea Group: the water dispenser line (GI-WD) and the dishwasher line (GI-D), as illustrated in Fig.~\ref{f4}-\ref{f5}.
The dataset statistics for each production line are detailed in Table~\ref{t1}.
All training and testing experiments were conducted on a workstation equipped with an Intel Xeon Gold 6342 CPU (2.80 GHz), 504 GB RAM, and an NVIDIA A800 80G GPU.
The final system was deployed on a device with an Intel Core i7-11700 CPU, 64 GB RAM, and an NVIDIA RTX 2080 GPU, enabling real-time inference in practical industrial settings.
Industrial videos were captured at 2560×1440 resolution using a DS-2CD3347DWDV3-L camera.
\subsection{Ablation Study}
\subsubsection{Generalization in Dataset Construction}
To evaluate the generalization capability of Grounding DINO in industrial dataset construction, we conducted comparative experiments against YOLOv5-General (YOLOv5-G) and YOLOv5-Overfit (YOLOv5-O). Grounding DINO and YOLOv5-G were trained on public datasets, whereas YOLOv5-O was trained on domain-specific industrial data.
\begin{table}[!htb]
	\scriptsize
	\centering
	\caption{Dataset scale of two manufacture lines.}
	\label{t1}
	\begin{tabular}{l@{\hskip 6pt}c@{\hskip 6pt}c@{\hskip 6pt}c@{\hskip 6pt}c@{\hskip 6pt}c@{\hskip 6pt}c}
		\toprule
		\multirow{2}{*}{Scenarios} & \multirow{2}{*}{Video Frame Count} & \multicolumn{5}{c}{Dataset Scale} \\
		\cmidrule(lr){3-7}
		& & $Act_1$ & $Act_2$ & $Act_3$ &$Act_4$& $NG$ \\
		\midrule
		GI-D & 3596 & 770 & 941 & 196 &--& 16179 \\
		GI-WD  & 3276 & 1500 & 1744 &2011 &1976 & 5864 \\
		\bottomrule
	\end{tabular}
\end{table}

\begin{table*}
	% \scriptsize  % 或 \tiny 
	% \setlength{\tabcolsep}{2pt} 
	\centering
	\caption{Accuracy and recall results under different thresholds. Bold font was the maximum recall value in each group.}
	\label{t2}
	\begin{tabular}{ccccc}
		\toprule
		\multicolumn{5}{c}{$T_{\text{text}} = 0.3$} \\
		\midrule
		Scenarios & Models & Hyperparameters & Accuracy & Recall \\
		\midrule
		\multirow{3}{*}{GI-WD} 
		& YOLOv5-G & Conf=0.1, NMS=0.9 & 90.14\%  & 75.84\% \\
		& YOLOv5-O & Conf=0.3, NMS=0.9 & 96.57\% & 86.41\% \\
		& Grounding DINO & Box=0.3, Text=(from 0.1 to 0.9) & 72.67\%  & \textbf{89.36\%} \\
		\midrule
		\multirow{3}{*}{GI-D} 
		& YOLOv5-G & Conf=0.1, NMS=0.1 & 91.71\%  & 70.92\% \\
		& YOLOv5-O & Conf=0.1, NMS=0.9 & 98.05\%  & 81.74\% \\
		& Grounding DINO & Box=0.4, Text=(from 0.1 to 0.9) & 83.19\%  & \textbf{83.99\%} \\
		\midrule
		\multicolumn{5}{c}{$T_{\text{text}} = 0.4$} \\
		\midrule
		Scenarios & Models & Hyperparameters & Accuracy & Recall \\
		\midrule
		\multirow{3}{*}{GI-WD} 
		& YOLOv5-G & Conf=0.3, NMS=0.2 & 92.92\%  & 71.18\% \\
		& YOLOv5-O & Conf=0.3, NMS=0.9 & 95.33\%  & 85.34\% \\
		& Grounding DINO & Box=0.3, Text=(from 0.1 to 0.9) & 67.61\%  & \textbf{87.32\%} \\
		\midrule
		\multirow{3}{*}{GI-D} 
		& YOLOv5-G & Conf=0.1, NMS=0.1 & 89.35\%  & 65.39\% \\
		& YOLOv5-O & Conf=0.3, NMS=0.9 & 93.60\%  & \textbf{75.49\%} \\
		& Grounding DINO & Box=0.4, Text=(from 0.1 to 0.9) & 81.44\%  & 71.07\% \\
		\bottomrule
	\end{tabular}
\end{table*}
\begin{figure*}
	\centering
	\centerline{\includegraphics[width=7in]{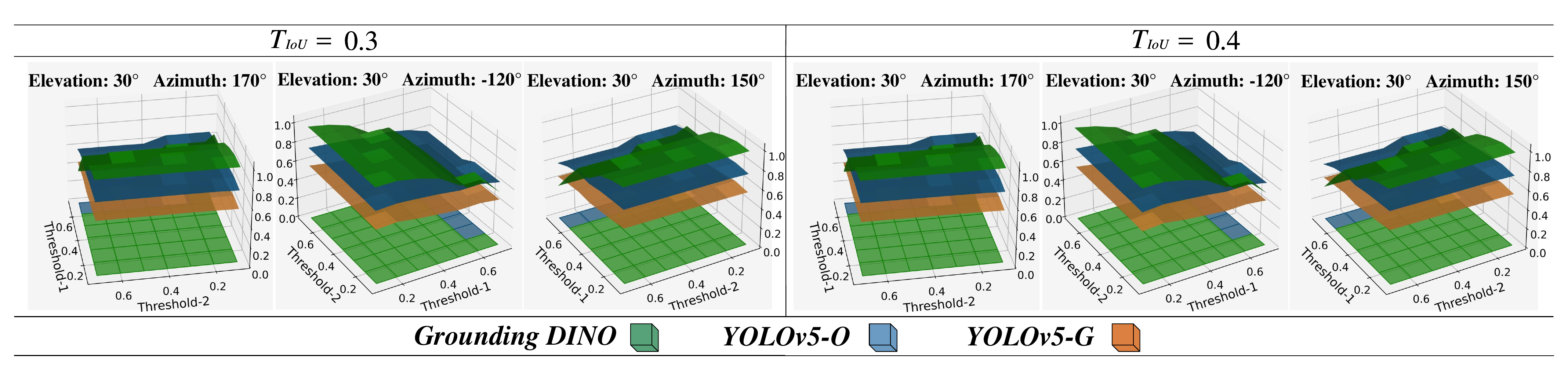}}
	\vspace{-6pt}
	\caption{Recall surfaces under all hyperparameter combinations are shown. In each plot, the vertical axis represents accuracy or recall, while the two horizontal axes span from 0.0 to 1.0. To ensure a comprehensive surface analysis, three observation angles (elevation and azimuth) were used for each surface comparison. To highlight the surface maximum value of the corresponding index for each model, we projected the model colors onto the Threshold 1-Threshold 2 plane. The surfaces in green, blue, and orange represent Grounding DINO, YOLOv5-O, and YOLOv5-G, respectively.}\label{f6}
\end{figure*}

Table~\ref{t2} showed results of the accuracy and recall of the three models on the different MLs. It was noticed that two tunable hyperparameters were contained in the Grounding DINO (i.e., Box and Text) and YOLO (i.e., Conf and NMS). 
The Grounding DINO was used to solve the low recall issues of YOLOv5-G in industrial scenes wherein specific actions were preferred to be detected as many as possible in order for IHAR systems to achieve high accuracy. 
In detection tasks, $T_{IoU}$ was usually selected between 0.3 and 0.4.
In fact, Table~\ref{t2} showed that the recall of the Grounding DINO was 13.52\% and 2.95\% higher on the GI-WD ML and 13.81\% and 2.25\% higher on the GI-D ML, over the YOLOv5-G and YOLOv5-O when $T_{IoU}$ was set to 0.3. 
Moreover, the recall of the Grounding DINO was 16.14\% and 1.39\% higher on the GI-WD ML over the YOLOv5-G and YOLOv5-O, with $T_{IoU}$ set to 0.4. 
Due to the fast handling actions in the GI-D ML, some images were blurred, and the positioning of the hand frames by the Grounding DINO had certain deviations, causing relatively poor recognition effect on the blurred scenarios. By adapting to various visual features and contexts, the Grounding DINO generalized its detection capabilities to unseen scenarios, facilitating its applications on object sifting and detecting and enhancing the robustness and reliability of the IHAR system. 

We also traversed all hyperparameter combinations to enable a more objective and fair assessment of the three models. As shown in Fig.~\ref{f6}, by testing on the MLs, different recall surfaces were generated. Specifically, the YOLOv5-G demonstrated noticeably lower accuracy over the YOLOv5-O, while the Grounding DINO exhibited comparable or higher accuracy when only some hyperparameter combinations were selected.
Moreover, Grounding DINO clearly demonstrated superior recall performance over the other two methods, indicating its strong generalization capabilities across different scenarios as a detection model. 
With suitable hyperparameters, the Grounding DINO could be directly applied to object sifting and detection of annotation systems, effectively reducing manual ACs and minimizing the risk of missing objects.

\subsubsection{Action Classification Accuracy via BLIP-2}
\begin{figure}
	\centering
	\centerline{\includegraphics[width=3.5in]{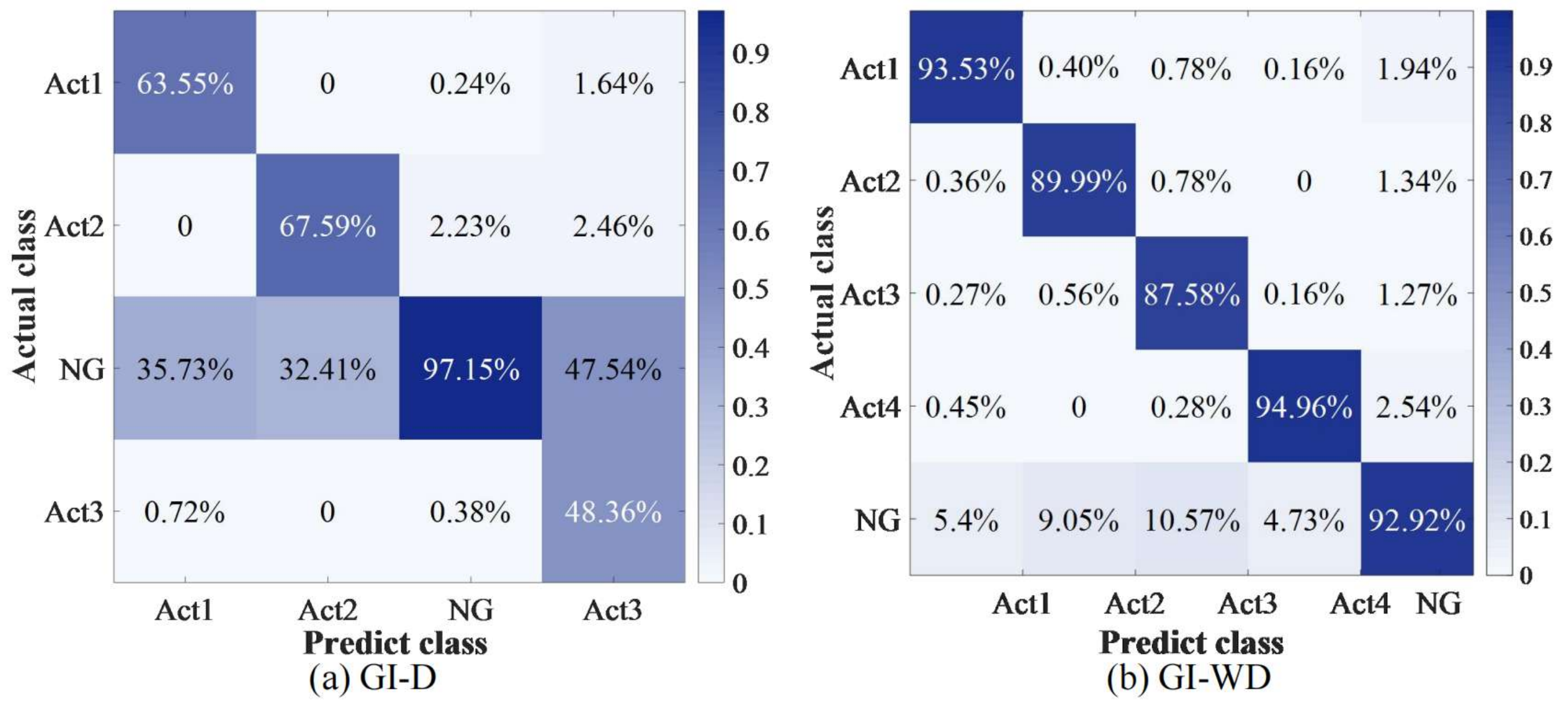}}
	\vspace{-6pt}
	\caption{The confusion matrix showing the percentages of correct predicted and false predicted action predictions using the large-scale model's image encoder in the following scenarios: GI-D, GI-WD.}\label{f7}
\end{figure}

We employed a frozen BLIP-2 image encoder to refine action classification based on the detected bounding boxes from Grounding DINO. As shown in the confusion matrix (Fig.~\ref{f7}), the classification accuracy was consistently high across datasets. Although the GI-D dataset exhibited slightly lower performance due to motion blur, these uncertain samples were filtered into an ``NG'' (No Good) class to avoid contaminating the final training set.

Across GI-D and GI-WD, we collected a total of 9,562 action instances: 8,950 were correctly classified, 25 were misclassified, and 587 were assigned to the NG class. The resulting average classification accuracy reached 99.72\%, demonstrating the effectiveness of our hybrid detection-classification pipeline. The use of the NG class notably reduced the need for manual intervention.
\subsubsection{ViT-L Fine-Tuning with LoRA }
\begin{figure}
	\centering
	\centerline{\includegraphics[width=3.5in]{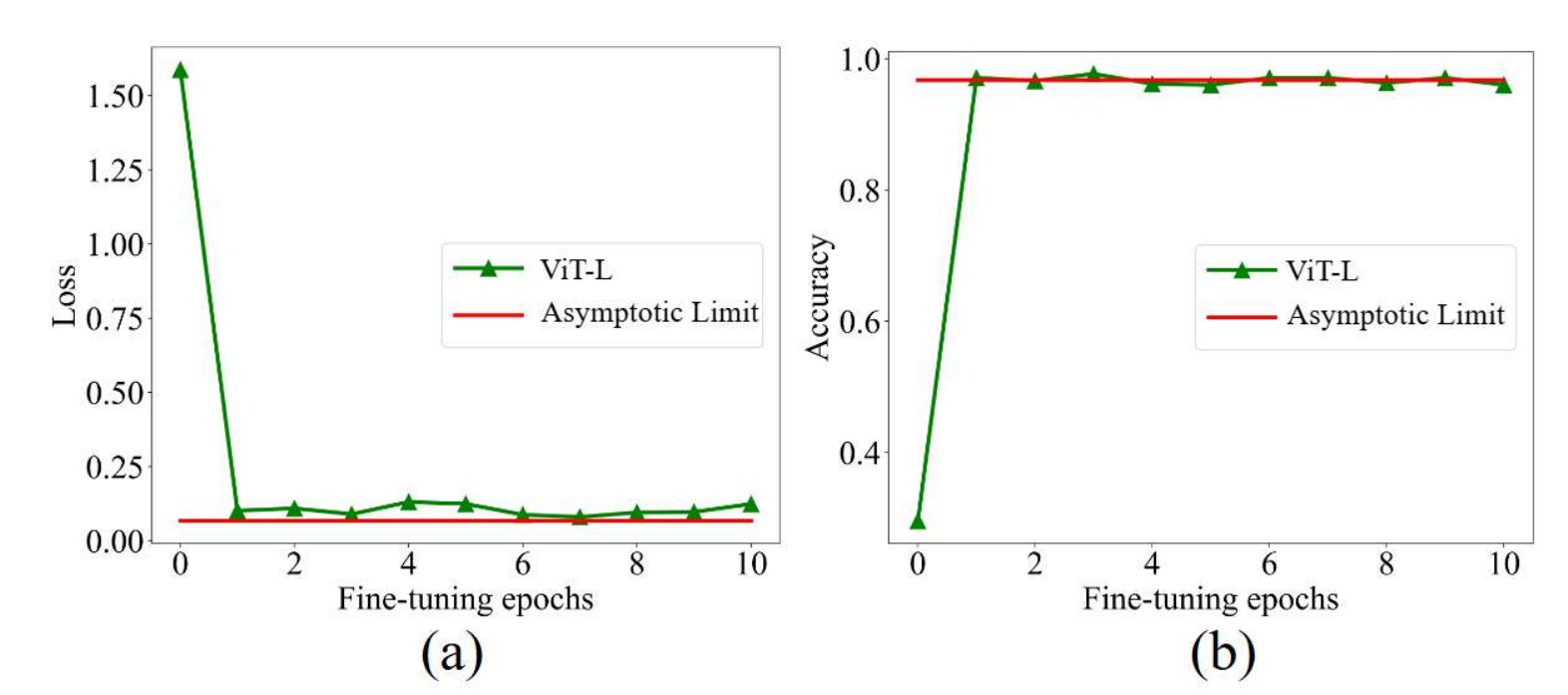}}
	\vspace{-6pt}
	\caption{The confusion matrix showing the percentages of correct predicted and false predicted action predictions using the large-scale model's image encoder in the following scenarios: GI-D, GI-WD.}\label{f8}
\end{figure}

To adapt the ViT-L model to industrial scenarios, we fine-tuned it using the following settings: rank $r=16$, learning rate $5 \times 10^{-3}$, and batch size 64. As illustrated in Fig.~\ref{f8}, the model rapidly converged within the first five epochs, achieving an average accuracy of 96.44\%. The use of LoRA effectively reduced the trainable parameters while mitigating overfitting, confirming its suitability for low-resource transfer learning.
\begin{table*}[!htb]
	\centering
	\caption{Comparisons of image annotation costs (ACs) and detection accuracy under two annotation methods, wherein ACs contained the number of images to be manually annotated (annotation workloadm, AW), and the time that was manually annotated (annotation time, AT). ImpV represented the improvement of LRIAR over conventional methods.}
	\label{t3}
	\begin{tabular}{c|c|c|c}
		\hline
		\multicolumn{4}{c}{Average ACs from GI-D and GI-WD Manufacturing Lines (MLs) } \\ \hline
		\multicolumn{1}{c|}{Sources of ACs} & \multicolumn{1}{c|}{Indices of costs} & \multicolumn{1}{c|}{Baseline method} & \multicolumn{1}{c}{Ours (ImpV)} \\\hline
		\multirow{1}{*}{\begin{tabular}[c]{@{}c@{}}Action Recognition\end{tabular}} & AW: total images (images for all 20 posts in a single ML) & 2,000,000 & \textbf{1,000,000 (-50\%)} \\ \hline
		& AW: detector training (images/post) & 500 & \textbf{100 (-80\%)} \\
		\begin{tabular}[c]{@{}c@{}}Detection\end{tabular} &	AT: detection (hours/post)& 4 & \textbf{0.8 (-80\%)} \\
		& AT: detection task (hours/ML)& 80 & \textbf{16 (-80\%)} \\ \hline
		& AW: classifier training (actions/post)& 200 & \textbf{20 (-90\%)} \\
		\begin{tabular}[c]{@{}c@{}}Classification\end{tabular} & AT: classifier training (hours/post) & 24 & \textbf{4 (-83.3\%)} \\
		& AT: classification (ML/post)& 480 & \textbf{80 (-83.3\%)} \\ \hline
		\multicolumn{2}{c}{\begin{tabular}[c]{@{}c@{}}Detection Accuracy \end{tabular}} &  \multicolumn{1}{|c|}{92.02\%} & \textbf{95.71\% (+3.69\%)} \\ \hline
	\end{tabular}
\end{table*}
\subsection{Cost Analysis of Dataset Construction}

Table~\ref{t3} presents the annotation costs (ACs) for both the conventional method (YOLOv5-O + ResNet-18) and LRIHAR, including deployment expenses for AR, detection, and classification.
In the conventional method, data annotation is performed manually, as is common in most industrial annotation tasks. In contrast, LRIHAR employs a LSFM-based annotation method, as described in Section \ref{3.1}.

For comparison of the two annotation methods, we utilized 2,000,000 raw images from inspection posts (IPs) processed by a single machine learning model (ML) per hour. The use of Grounding DINO in LRIHAR effectively filtered out 1,000,000 images, achieving a high recall rate with a significant level of accuracy. This reduced the overall annotation workload (AW) and automatically categorized images, without human input, into the NG class.

LRIHAR leveraged Grounding DINO to automatically select 500 training images, with only 100 requiring correction for detection. Compared to YOLOv5-G, the AW in LRIHAR decreased by 80\%. 
Furthermore, the process consumed just 0.8 hours per IP, and 16 hours for 20 IPs, marking an 80\% improvement over YOLOv5-G. Additionally, the LSFM-based annotation method in LRIHAR also reduced the ACs for classification. As shown in Table~\ref{t3}, ResNet-18 required 200 normal images per target action category, similar to the training for detection. However, only 20 images out of these 200 needed manual correction, reducing the time spent to 4 hours, which is one-sixth of the time required for manual annotation. Expanding this process to 20 IPs resulted in a total time of 80 hours, which is 83.3\% shorter than the 480 hours required for manual annotation. Interestingly, when the annotation system was deployed in LRIHAR, the detection task's accuracy improved by 3.69\% over YOLOv5-O. These reductions in ACs can be attributed to the strong generalization capabilities of both Grounding DINO and BLIP-2 used in LRIHAR, enabling the replacement of human workers in handling the alignment task between annotation and image semantics in industrial environments.
\subsection{Real-Time Classification Performance Comparison}
Table~\ref{t4} summarizes the deployment performance of ResNet-18, ViT-L, and our LRIAR method. LRIAR achieved the highest average classification accuracy (95.71\%) among all tested models. However, the model exhibited slightly higher inference latency, attributed to the use of the ViT backbone.
Despite this, LRIAR demonstrated a favorable trade-off between recognition accuracy and deployment cost, offering strong practical value for industrial real-time action recognition systems.
\begin{table}[!htbp]
	\scriptsize
	\setlength{\tabcolsep}{2pt}
	\centering
	\caption{Comparisons of accuracy and time-consuming with various classifiers. Bold font represented the highest accuracy in the same ML scenario.}
	\label{t4}
	\begin{tabular}{ccccc}
		\hline
		\multicolumn{5}{c}{Accuracy comparisons of various classifiers for single ML} \\ \hline
		&ResNet-18 (General)&ResNet-18 (Re-trained)& VIT-L& Ours \\ \hline
		GI-D & 92.94\% & 93.13\% & 94.41\% & \textbf{95.92\%}  \\
		GI-WD & 91.10\% & 94.15\% & 94.82\% & \textbf{95.75\%} \\
		Average & 92.02\% & 93.64\% & 94.62\% & \textbf{95.71\%}\\ \hline
	\end{tabular}
\end{table}
\section{Conclution}
Action recognition in large-scale manufacturing faces persistent challenges in balancing annotation cost, real-time performance, and recognition accuracy. This paper presents LRIAR, a lightweight real-time framework that leverages large-scale foundation models to address these issues comprehensively.
We construct an industrial-scale action dataset by integrating Grounding DINO and BLIP-2 for automatic annotation, significantly reducing manual labeling effort. Compared to YOLOv5, LRIAR improves average recall by 6.4\%. For classification, ViT-L is fine-tuned with LoRA, achieving 96.44\% accuracy. The framework further ensures real-time inference and generalization across diverse manufacturing lines.
Extensive evaluations demonstrate that LRIAR not only achieves high recognition performance but also reduces annotation and deployment costs, offering a scalable and practical solution for intelligent action understanding in smart manufacturing.
\section*{Acknowledgment}
This work was supported in part by the Hunan Key Scientific and Technological Research Foundation under Grant 2024QK2001, in part by the Guangdong Basic and Applied Basic Research Foundation under Grant 2022A1515140121, and in part by the Initiation Funding of Foshan Graduate School of Innovation, Northeastern University under Grant 200076421002.

\bibliographystyle{cas-model2-names}
%\bibliographystyle{unsrt}
\bibliography{ref01}

% % Here goes the biography details.

% % Here goes the biography details.

\end{document}